%% file: data_compression_with_bayesian_attention_networks.tex
\documentclass{article} 
\usepackage{iclr2021_conference,times}

\input{math_commands.tex}

\usepackage{amssymb}
\usepackage{hyperref}
\usepackage{url}
\usepackage{algorithm2e}

\title{Bayesian Attention Networks for Data Compression}

\author{Michael Tetelman \\
Volkswagen Group of America Innovation Center California, BayzAI\\
\texttt{michael.tetelman@gmail.com}
}

\iclrfinalcopy 
\begin{document}

\maketitle
\begin{abstract}
The lossless data compression algorithm based on Bayesian Attention Networks is derived from first principles. Bayesian Attention Networks are defined by introducing an attention factor per a training sample loss as a function of two sample inputs, from training sample and prediction sample. By using a sharpened Jensen's inequality we show that the attention factor is completely defined by a correlation function of the two samples w.r.t. the model weights. Due to the attention factor the solution for a prediction sample is mostly defined by a few training samples that are correlated with the prediction sample. Finding a specific solution per prediction sample couples together the training and the prediction. To make the approach practical we introduce a latent space to map each prediction sample to a latent space and learn all possible solutions as a function of the latent space along with learning attention as a function of the latent space and a training sample. The latent space plays a role of the context representation with a prediction sample defining a context and a learned context dependent solution used for the prediction.
\end{abstract}
\section{Introduction: Lossless Compression as a Prediction Problem}

Data compression is a very important area of study for both practical applications as well as for fundamental research in information theory \citep{salomon2006},\citep{mackay2002information}, \citep{matt_mahoney_dcbook}, \citep{wiki_LosslessCompression} and subject of many competitions, like  \citep{wiki_Hutter_Prize}. See also \citep{matt_mahoney_dc}. The neural compression is an active research area \citep{kingma2019bitswap}.

The fundamental approach to compressing a sequence of data elements is to use a probabilistic model that is able to reproducibly compute a probability of the sequence via conditional probabilities of all elements and then encode the sequence probability with an entropy encoder \citep{shannon1948},\citep{rissanen1976arithmetic},\citep{pasco1976}.

The direct consequence of that approach is that the finding of a good predictive model becomes  critical for the development of the high performance compression algorithm.

For a sequence of elements 
\begin{equation}
s = a_1 a_2 a_3 .. a_n.. a_N
\end{equation}
the probability of the sequence is a product of conditional probabilities of each element
\begin{equation}
P(s) = P(a_1)P(a_2|a_1)P(a_3|a_1a_2)..P(a_n|s_n)..P(a_N|s_N),
\end{equation}
where sub-sequences $s_n$ are defined as $s_n=a_1 a_2 .. a_{n-1}, n=2..N$.
To simplify we will consider conditional dependencies for all probabilities as  functions of fixed-length sequences $s_{n,n+l}$ of length $l$ with left-side zero padding as needed. Then the probability of the sequence $s$ will be
\begin{equation}
P(s) = \prod_{n=1}^N P(a_n|s_{n-l,n}), \; s_{n-l,n} = a_{n-l}a_{n-l+1}..a_{n-1}.
\end{equation}
By defining a model $P(a|s, w)$ for probability of an element $a$ given a previous sequence  $s$ and parametrized by weights $w$ we obtain a parametrized probability of any sequence $s$
\begin{equation}
P(s|w) = \prod_{n=1}^N P(a_n|s_{n-l,n},w), \; s_{n-l,n} = a_{n-l}a_{n-l+1}..a_{n-1}.
\end{equation}
For a known training dataset of independent sequences $\{s_i, i=1..M\}$ the probability of the data for a selected model is given by Bayesian integral over weights as it follows from the Bayes theorem \citep{bayes1763}
\begin{equation}
Prob(\{s_i\}|H) = \int_w \prod_{i=1}^M P(s_i|w) P_0(w|H) dw,
\end{equation}
where $P_0(w|H)$ is a prior of weights conditional on hyperparameters $H$.

The prediction of the probability $P(s|\{s_i\},H)$ of a new sequence $s$ then will be given by an average
\begin{equation}
\label{P(s)}
P(s|\{s_i\},H) = \langle P(s|w)\rangle = \int_w P(s|w) \prod_{i=1}^M P(s_i|w) P_0(w|H) dw \;/\; Prob(\{s_i\}|H).
\end{equation}
We will call the new sequence $s$ a test sequence.

The prediction probability $P(s|\{s_i\},H)$ is an average over weights with the most significant contribution coming from training sequences that maximize both corresponding probabilities of training data and the test sequence $P(s|w)$ for the same weights in the Eq.(\ref{P(s)}). The resulting average probability is high when the test sequence $s$ is similar to the most typical training sequences.
However, the average will be low when $s$ is very different from the typical training data.
The problem is that all training sequences contributing equally to the average and a good prediction is possible only for a test sequence that is very similar to training data.

To improve the prediction of the probability for the given sequence $s$ we will introduce the importance weights for training data samples in the Eq.(\ref{P(s)}) that will increase contributions of samples with sequences $s_i$ similar to the test sequence $s$ and reduce contributions of other samples.

The idea is an extension of a very well known approach used for compression that maps first a prediction sample to a certain context and then a prediction for the sample is made using a context specific predictor.

The approach with importance weights described in this paper leads to a definition of Bayesian Attention Networks (BAN) considered next.

The idea to account for long-range correlations between data samples for achieving better compression has a long history. The attention mechanism recently proposed in \citep{vaswani2017attention} was a great success for sequence prediction tasks as well as for other problems including image recognition \citep{dosovitskiy2020image}.

\section{Prediction of Sequence Probabilities with Bayesian Attention Networks (BAN)}
We will define BAN in a few steps. First, we introduce the importance factors $\rho_i(s)$ for sequences by modifying contribution of each training data sample in the Bayesian integrals
\begin{eqnarray}
\label{BAN}
P(s|\{s_i\},H) &=& \int_w P(s|w) \prod_{i=1}^M e^{-\rho_i(s) l(s_i|w)} P_0(w|H) dw \;/\; Prob(\{s_i\}|H),\\
Prob(\{s_i\}|H) &=& \int_w \prod_{i=1}^M e^{-\rho_i(s) l(s_i|w)} P_0(w|H) dw.
\end{eqnarray}
Here $l(s_i|w)$ are loss functions for sequences defined as logs of probabilities $l(s|w)=-\log(P(s|w))$. The importance factors $\rho_i(s)$ are assigned to each training sequence $s_i$ and are functions of predicted sequence $s$. The importance factors must be constrained by normalization condition $\sum_i\rho_i(s)=M$, so the total contribution of all training sequences is preserved.

The meaning of the importance factors is very simple - it defines the importance of a given training sample $s_i$ to influence a solution for the prediction probability of the test sequence $s$ with higher importance $\rho_i(s)$ increasing contribution of the sample $s_i$ to predicting $s$. 

In the limit when unnormalized importance is only zero or one it is equivalent to clustering when the solution for predicting sample $s$ is defined by a few training samples $s_i$ correlated with the prediction sample.

We will clarify the definition of importance weights in the following steps.

Next, we will approximate the average of probability of a sequence as product of averages for predictions of probabilities of elements conditional on corresponding immediate sub-sequences
\begin{equation}
P(s) = P(a_1a_2..a_N) = \langle P(a_1|s_1,w)\rangle \langle P(a_2|s_2,w)\rangle .. \langle P(a_N|s_N,w)\rangle.
\end{equation}
Then the prediction of the probability of an element is given by
\begin{equation}
\label{BANa}
\langle P(a|s,w)\rangle = \int_w P(a|s,w)\prod_{i=1}^Ke^{-\rho_i(s)l(a_i|s_i,w)}P_0(w|H)dw\; /\; Prob(\{a_i\}|H)
\end{equation}
with the normalization factor $Prob(\{a_i\}|H)$ that keeps $\langle P(a|s,w)\rangle$ normalized defined as follows
\begin{equation}
\label{BANb}
Prob(\{a_i\}|H) = \int_w \prod_{i=1}^Ke^{-\rho_i(s)l(a_i|s_i,w)}P_0(w|H)dw.
\end{equation}
Here the importance factors depend only on input sequences in each sample and not on predicted elements.
The index $i$ runs over all $K$ sequence elements $a_i$ in training data in Eqs.(\ref{BANa},\ref{BANb}). The losses $l(a_i|s_i,w)$ are per element in training sequences and the importance factors $\rho_i(s)$ control the contributions of each training sample $i$ and depend on the sub-sequence $s$ of the element $a$ in the prediction probability $P(a|s)$. The importance factors are normalized by condition $\sum_i\rho_i(s)=K$ where $K$ is a total number of training data samples.

The Eqs.(\ref{BANa},\ref{BANb}) are defining the Bayesian Attention Network with attention variable $\rho_i(s)$ which is a function of two sequences. The attention or importance factors allow to take into account long range interactions between different data samples.

It is important to note that for any approximation of the integrals in Eqs.(\ref{BANa},\ref{BANb}) the solution is dependent on the input sequence $s$ in the data sample we are making prediction for $P(a|s)$, which requires computing the integrals for every test sample.

To resolve this problem and find the practical computing approach we will use a latent variable $z$ representation with the encoder-decoder network that is considered in the next section. 

\section{Computing Importance via Extended Variational Approximation}
Let's define encoder $\rho(z|s)$ and decoder $\rho(s|z)$. To do a prediction for a sample (a,s) the encoder $\rho(z|s)$ gives a probability of $z$ for input $s$ and the decoder $\rho(s_i|z)$ gives a probability of a training sequence $s_i$ for a given $z$.

The decoder allows to define the normalized importance factor 
\begin{equation}
\rho_i(z)=K\rho(s_i|z)/\sum_j\rho(s_j|z).
\end{equation}
With that we can reformulate Eqs.(\ref{BANa},\ref{BANb}) as follows
\begin{eqnarray}
\label{BANz}
\langle P(a|s,w)\rangle &=& \int_{w,z} P(a|s,w)\prod_{i=1}^Ke^{-\rho_i(z)l(a_i|s_i,w)}P_0(w|H)\rho(z|s)dwdz\; /\; Prob(\{a_i\}|H)\nonumber\\
Prob(\{a_i\}|H) &=& \int_{w,z} \prod_{i=1}^Ke^{-\rho_i(z)l(a_i|s_i,w)}P_0(w|H)\rho(z|s)dwdz
\end{eqnarray}
Due to Bayesian theorem \citep{bayes1763} the encoder probability related to the decoder probability and  can be expressed via decoder probability and priors for $z$ and $s$
\begin{equation}
\rho(z|s) = \frac{\rho(s|z)\rho(z)}{\rho(s)}.
\end{equation}
By introducing an approximation for the encoder $q(z|s)$ we can use a variational method for simplifying the integral over $z$ following \citep{kingma2014autoencoding}
\begin{equation}
\label{var}
\int_z e^{-L(z)}\rho(s|z)\rho(z)dz \geqslant \exp\left(-\int_z q(z|s)L(z,w)dz + \int_z q(z|s)\log\frac{\rho(s|z)\rho(z)}{q(z|s)}dz\right),
\end{equation}
where total training loss is $L(z,w)=\sum_i \rho_i(z)l(a_i|s_i,w)$. Maximizing the right side of Eq.(\ref{var}) w.r.t. parameters of $q(z|s)$ allows to find the encoder $q$.

However, the variational approximation that is based on Jensen's inequality,  \citep{jensen1906} $\langle\exp(x)\rangle\geqslant\exp(\langle x\rangle)$, results in the loss of dependency on importance factors in the prediction probability in the Eqs.(\ref{BANz}).
To retain that dependency we have to use a sharpened Jensen's inequality, (see Appendix A for the proof)
\begin{equation}
\label{sharpJensen}
\langle e^{x}\rangle \geqslant e^{\langle x\rangle + \frac{1}{2}\langle (x-\langle x\rangle)^2\rangle}.
\end{equation}
Then the probability of prediction can be maximized w.r.t. importance weights.
Now we can find that the variation of the log of the prediction probability is equal to a correlation function of test and train losses
\begin{eqnarray}
\frac{\delta\log{\langle P(a|s,w)\rangle}}{\delta\rho_i(z)} = \langle l(a_i|s_i,w) l(a|s,w)\rangle_w-\langle l(a_i|s_i,w)\rangle_w\langle l(a|s,w)\rangle_w \approx \nonumber \\ \sum_j \sigma_j^2 \frac{\partial l(a_i|s_i,w)}{\partial w_j} \frac{\partial l(a|s,w)}{\partial w_j},
\end{eqnarray}
where averages $\langle(..)\rangle_w$ are computed with a posterior distribution  of weights 
\begin{equation}
\label{post}
Post(w|H_0,z) = \prod_{i=1}^Ke^{-\rho_i(z)l(a_i|s_i,w)}P_0(w|H_0),
\end{equation}
by expanding the weights around the means up to the second order and averaging over it.

\section{Complete Loss and Recurrent Update Rules for Hyperparameters}
Due to the sharpened Jensen's inequality in Eq.(\ref{sharpJensen}) the complete loss includes a cross-correlation term between the test sample loss and training sample losses
\begin{equation}
\label{loss}
\int_{w,z}P_0(w|H_0) q(z|s)\left(L(z,w) - l(a|s,w)L(z,w)\right)dwdz - \int_z q(z|s)\log\frac{\rho(s|z)\rho(z)}{q(z|s)}dz,
\end{equation}
where the training loss $L(z,w)=\sum_i \rho_i(z)l(a_i|s_i,w)$.

Finding the importance factors by maximizing the prediction probability for test sample requires to minimize only the cross-correlation loss due to normalization factor in Eq.(\ref{BANz}). All other parameters could be found by minimizing the loss in the Eq.(\ref{loss}).

Let's use a posterior distribution for weights that depends on initial hyperparameters $H_0$ and the latent variable $z$ in the Eq.(\ref{post}).
The method developed in \citep{tetelman2020compression} allows to compute approximations of the Bayesian integrals by re-parametrizing a prior of weights $P_0(w|H)$ to represent a posterior
\begin{equation}
Post(w|H',z) = \prod_{i=1}^Ke^{-\rho_i(z)l(a_i|s_i,w)}P_0(w|H_0) \sim P_0(w|H(z,H_0)).
\end{equation}
The re-parametrization is computed recursively. With Gaussian priors for all network weights and Gaussian encoder $q(z|s)$ we can find the following update rules for hyperparameters: (mean, variance) pairs $(\mu,\sigma^2)$ for all weights of all networks. The training update steps are as follows
\begin{eqnarray}
i &\sim& \mathbf{Uniform}[1..K] \\
z(s,v) &\sim& q(z|s,v) \\
w(z,v) &\sim& P_0(w|\mu_w(z|v),\sigma_w(z|v)) \\
\rho_i(z,u) &=& K\rho(s_i|z,u)/\sum_j\rho(s_j|z,u) \\
\mathbf{grad_v} &=& \frac{\partial}{\partial v} \bigg[\rho_i(z,u)l(a_i|s_i,w(z,v))-\log{\frac{{•}\rho(s|z,u)\rho(z)}{q(z|s,v)}}\bigg] \\
\Delta\mu_v &+=& -\varepsilon\sigma_v^2 \mathbf{grad_v} \\
\Delta\frac{1}{\sigma_v^2} &+=& \varepsilon (\mathbf{grad_v})^2 \\
\mathbf{grad_u} &=& \sum_j \bigg(\frac{\sigma_{w_j}^2}{K} \frac{\partial l(a_i|s_i,w)}{\partial w_j} \frac{\partial l(a|s,w)}{\partial w_j}\bigg) \frac{\partial\rho_i(z,u)}{\partial u} \\
\Delta\mu_u &+=& -\varepsilon\sigma_u^2 \mathbf{grad_u} \\
\Delta\frac{1}{\sigma_u^2} &+=& \varepsilon (\mathbf{grad_u})^2.
\end{eqnarray}
Here the learning rate $\varepsilon$ is equal to the inverse number of epochs $T$: $\varepsilon=1/T$.

The prediction for a sample with input $s$ is done by sampling $z\sim q(z|s,v)$, computing a mean for weight $w$ equal to $\mu_w(z,v)$ and finally computing a prediction probability $P(a|s,\mu_w)$.

\section{Conclusion}

The compression is defined by computing a probability of a prediction sample that is then encoded with the entropy coder. The contribution of this paper is to derive the compression algorithm by defining the Bayesian Attention Networks for predicting a probability of a new sample by introducing an attention or importance factor per the training sample that is dependent on the prediction sample. The attention factors allow to find a specific solution for each prediction sample that is influenced by a few training samples most correlated with the prediction sample. 

Because the solution becomes specific to a given prediction sample the training and the prediction processes are coupled together. To make this approach practical we introduce a latent space representation with encoder-decoder networks that allows to separate training and prediction by mapping any prediction sample to a latent space and learning all solutions as a function of the latent space along with learning attention as a function of the latent space and a training sample.

The latent space plays a role of the context representation with a prediction sample defining a context and a learned context dependent solution used for the prediction.

\bibliography{data_compression_with_bayesian_attention_networks}
\bibliographystyle{iclr2021_conference}

\appendix
\section{Appendix}

To prove the sharpened Jensen's inequality we note that for a normal distribution 
\begin{equation}
\langle e^x\rangle = e^{\langle x\rangle + \frac{1}{2}\langle (x-\langle x\rangle)^2\rangle}
\end{equation}
As Gaussian mixture is an universal approximator for any (non-pathological) distribution, see \citep{DLbook2016} p.65, then it directly proves the sharpened Jensen's inequality in Eq.(\ref{sharpJensen}):
\begin{equation}
\int_x dx \sum_k A_k G(x|\mu,\sigma) e^x = \sum_k A_k e^{\mu_k+\frac{1}{2}\sigma_k^2} \geqslant e^{\mu + \frac{1}{2}\sigma^2}.
\end{equation}
Here $\sum_k A_k = 1$ and $\mu=\sum_k A_k \mu_k, \sigma^2=\sum_k A_k \sigma_k^2$.

\end{document}

%% file: math_commands.tex
\usepackage{amsmath,amsfonts,bm}

\def\eqref#1{equation~\ref{#1}}

\def\1{\bm{1}}

\DeclareMathAlphabet{\mathsfit}{\encodingdefault}{\sfdefault}{m}{sl}
\SetMathAlphabet{\mathsfit}{bold}{\encodingdefault}{\sfdefault}{bx}{n}

